\documentclass[final, 3p, 10pt]{elsarticle}

\usepackage{graphics}
\usepackage{graphicx}
\usepackage{amsfonts}
\usepackage{amsmath}
\usepackage{amssymb}
\usepackage{dsfont}
\usepackage{color}
\usepackage{hyperref}
\usepackage{subcaption}
\usepackage{epsfig}
\usepackage{float}
\usepackage{color}
\usepackage[ruled]{algorithm2e}
\usepackage{algorithmic}
\usepackage{natbib,hyperref}
\usepackage{lineno,hyperref}

\newcommand{\bx}{\mathbf{x}}

\newcommand{\bh}{\mathbf{h}}
\newcommand{\bu}{\mathbf{u}}

\newcommand{\bw}{\mathbf{w}}

\newcommand{\bm}{\mathbf{m}}

\newcommand{\bI}{\mathbf{I}}

\newcommand{\bV}{\mathbf{V}}
\newcommand{\bU}{\mathbf{U}}

\newcommand{\bW}{\mathbf{W}}

\newcommand{\bF}{\mathbf{F}}

\newcommand{\bZ}{\mathbf{Z}}
\newcommand{\bbell}{\boldsymbol{\ell}}

\newcommand{\bSigma}{\boldsymbol{\Sigma}}

\newcommand{\btheta}{\boldsymbol{\theta}}

\newcommand{\bphi}{\boldsymbol{\phi}}

\newcommand{\calG}{\mathcal{G}}

\newcommand{\calN}{\mathcal{N}}
\newcommand{\calP}{\mathcal{P}}

\newcommand{\calM}{\mathcal{M}}

\newcommand{\calU}{\mathcal{U}}
\newcommand{\calX}{\mathcal{X}}

\newcommand{\calV}{\mathcal{V}}

\newcommand{\calS}{\mathcal{S}}
\newcommand{\calZ}{\mathcal{Z}}

\newcommand{\R}{\mathbb{R}}

\newcommand{\bell}{\mathbf{\ell}}

\journal{Journal of Computational Physics}

\begin{document}

\begin{frontmatter}

\title{Bayesian learning of orthogonal embeddings for multi-fidelity Gaussian Processes}

\author[address1]{Panagiotis Tsilifis\corref{mycorrespondingauthor}}
\ead{panagiotis.tsilifis@ge.com}


\author[address1]{Piyush Pandita}
\ead{piyush.pandita@ge.com}

\author[address1]{Sayan Ghosh}
\ead{sayan.ghosh1@ge.com}

\author[addr2]{Valeria Andreoli}
\ead{valeria.andreoli@ge.com}

\author[addr2]{Thomas Vandeputte}
\ead{vandeput@ge.com}

\author[address1]{Liping Wang}
\ead{wangli@ge.com}

\address[address1]{Probabilistic Design \& Optimization Group, General Electric Research, Niskayuna, NY 12309, USA}
\address[addr2]{Aerodynamics \& Computational Fluid Dynamics Group, General Electric Research, Niskayuna, NY 12309, USA}

\cortext[mycorrespondingauthor]{Corresponding author}

\begin{abstract}
Uncertainty propagation in complex engineering systems often poses significant computational challenges related to modeling and quantifying probability distributions of model outputs, as those emerge as the result of various sources of uncertainty that are inherent in the system under investigation. 
Gaussian Processes regression (GPs) is a robust meta-modeling technique that allows for fast model prediction and exploration of response surfaces. 
Multi-fidelity variations of GPs further leverage information from cheap and low fidelity model simulations in order to improve their predictive performance on the high fidelity model. 
In order to cope with the high volume of data required to train GPs in high dimensional design spaces, a common practice is to introduce latent design variables that are typically projections of the original input space to a lower dimensional subspace, and therefore substitute the problem of learning the initial high dimensional mapping, with that of training a GP on a low dimensional space. 
In this paper, we present a Bayesian approach to identify optimal transformations that map the input points to low dimensional latent variables. 
The ``projection" mapping consists of an orthonormal matrix that is considered a priori unknown and needs to be inferred jointly with the GP parameters, conditioned on the available training data. 
The proposed Bayesian inference scheme relies on a two-step iterative algorithm that samples from the marginal posteriors of the GP parameters and the projection matrix respectively, both using Markov Chain Monte Carlo (MCMC) sampling. In order to take into account the orthogonality constraints imposed on the orthonormal projection matrix, a Geodesic Monte Carlo sampling algorithm is employed, that is suitable for exploiting probability measures on manifolds.
We extend the proposed framework to multi-fidelity models using GPs including the scenarios of training multiple outputs together.
We validate our framework on three synthetic problems with a known lower-dimensional subspace.
The benefits of our proposed framework, are illustrated on the computationally challenging three-dimensional aerodynamic optimization of a last-stage blade for an industrial gas turbine, where we study the effect of an 85-dimensional airfoil shape parameterization on two output quantities of interest, specifically on the aerodynamic efficiency and the degree of reaction.
\end{abstract}


\end{frontmatter}


\section{Introduction}
\label{sec:intro}

In addressing reliability and design optimization challenges in modern engineering and manufacturing processes, the need for an increasing accuracy in the predictive capabilities of the analysis codes has been of paramount importance. Taking into account the presence of aleatoric uncertainties resulting from the severe complexity of the physical system under investigation and the limited knowledge available to the experimenter, simulation-based uncertainty quantification (UQ) tasks can result in computational costs that are unacceptable in practice \cite{smith_uq, lemaitre, koch_size}. 

Metamodeling techniques \cite{wang_review} have gained increasing popularity in recent years, as they provide a way to alleviate the computational burden associated with quantifying uncertainties in computer model outputs and inverse modeling \cite{morokoff, tarantola}. The key idea is to replace the expensive computer solver with a surrogate that is cheap to evaluate and at the same time it maintains high predictive accuracy. These metamodels can consist of functional representations of the forward solver such as Polynomial Chaos \cite{ghanem_spanos, lemaitre_knio, xiu_karniadakis}, kernel-based methods \cite{cristianini, cortes}, relevance vector machines \cite{bilionis_rvm, tsilifis_rvm} and neural networks \cite{tripathy_nn, zhu} or they can be nonparametric models such as Gaussian Processes \cite{bilionis_mogp, chen_jcp, perdikaris_scicomp}. 

Gaussian Process (GP) regression techniques allow us to quantify the epistemic uncertainty associated with the limited number or training data points by providing explicit Bayesian posterior updates on the error bars \cite{ohagan, ohagan_quad}. Recent works have demonstrated their wide applicability for UQ analysis and probabilistic machine learning and have been exhaustively used for predicting solutions to differential equations, including local adaptations for detecting discontinuities \cite{bilionis_local}, discovering governing equations \cite{raissi_gp} and constructing latent variable models for solving inverse problems \cite{atkinson}. Challenges associated with efficiently training an accurate GP model, include acquiring the training dataset necessary to guarantee the desired predictive performance and tuning the model's hyperparameters \cite{rasmussen}. Collecting the data often requires a significant number of model simulations that increases exponentially as a function of the dimensionality of the input parameters and can easily become prohibitive due to computational budget limitations. On the other hand, training GPs using massive datasets, when available, poses numerical challenges that can result in poor performance. These are mainly associated with repeatedly performing covariance matrix inversions that can be extremely inaccurate and computer memory-demanding. Applying sparse techniques \cite{herbrich} and batch optimization \cite{hensman} have only partially managed to address these issues. 

An alternative approach involves leveraging a cheaper computer solver that provides with low-cost, yet less accurate observations to be used for training the GP, leading to formulations where the observable depends on more that one covariates. Such approaches, although already known to the geostatistics community as co-kriging \cite{cressie}, were formalized within the context of multi-fidelity simulations in the pioneering work of Kennedy \& O'Hagan \cite{kennedyohagan}. The auto-regressive scheme presented therein, decomposes the expensive simulation code as a sum of its cheap approximation and a discrepancy term, both modeled as independent Gaussian Processes. Techniques for training the model have been proposed \cite{forrester} and involve learning the degree of correlation between the codes, while a model calibration variation of the scheme is also available \cite{kennedyohagan_calib}. Similarly to standard GP regression in single-fidelity settings, training the auto-regressive scheme includes inverting large, ill-conditioned covariance matrices, leading to numerical inaccuracies. Le Gratiet \& Garnier \cite{le_gratiet_doe} offered an elegant solution to the problem that consists of decoupling the information stemming from different levels of fidelity and thus simplifying the covariance matrix structure. These ideas found wide applicability in discovering high dimensional response surfaces arising from dynamical systems \cite{paris_rspa}, model inversion \cite{paris_rsif}, multi-fidelity Bayesian optimization \cite{pang} and uncertainty propagation in the big-data regime \cite{paris_siam}.

From a different perspective, an efficient strategy for constructing a GP metamodel would involve addressing input dimensionality by means of reducing the number of variables in the input design space. Standard ways of doing so include sensitivity analysis \cite{saltelli} and unsupervised learning methods that explore correlations of the input variables, such as PCA \cite{pearson}, kernel PCA \cite{ma} or even truncating Karhunen-Lo\`eve expansions \cite{ghanem_spanos}. More sophisticated approaches to dimensionality reduction in the context of GPs include applying embeddings on the input design space in order to obtain a low dimensional latent variable space. Several criteria for discovering linear embeddings have been proposed in the literature. Constantine in his seminal work \cite{constantine_AS, constantine_book}, used a gradient-based approach to exploit what is termed Active Subscape (AS). The key step in this approach is to map the input space using a matrix derived by an orthogonal decomposition of the covariance of the gradient vector of the observable quantity. Intuitively the approach is sound as it explores the directions along which the output quantity exhibits most of its variability. The major limitation however is that it is impractical in black-box simulations where gradient information is hardly available. Gradient-free approaches that learn orthogonal \cite{tripathy_bilionis} or arbitrary \cite{garnett} linear mappings on-the-fly, have also been developed. In these works, the matrices were inferred using MLE or the Laplace approximation of the posterior of the linear mapping, respectively. At last, attempts to reduce the dimensionality using nonlinear mappings have been successfully applied for Bayesian optimization \cite{moriconi}, where the forward mapping to the latent variables is modeled using a feed-forward neural net while the original input reconstruction mapping is carried out using multi-output GPS. 

The ultimate goal in this work is to combine dimensionality reduction principles with multi-fidelity simulations so as to manifest their benefits in a single context. We seek to simultaneously address numerical issues and computational limitations arising in the presence of high input dimensionality and expensive data acquisition procedures. We thus present a unified framework where the classic auto-regressive Gaussian Process (ARGP) scheme can be built on a linearly embedded subspace that can be learned while training the model. To the authors best knowledge, such an approach is presented for the first time. Specifically, we build on the Kennedy-O'Hagan autoregessive scheme for modeling expensive computer outputs using a multi-fidelity source of information. Regardless of the quality of their output as an approximation of the high fidelity solver, we assume that the different fidelity codes are highly correlated and thus they exhibit similar dependence on the input variables, thus intuitively it is natural to assume that a common linear embedding can be applied to all GPs at different levels of fidelity. A training procedure for tuning the hyperparameters of the model is presented where the ARGP parameters are estimated using either MLE or Bayesian methods and the orthogonal embedding is inferred using a fully Bayesian approach that relies on the Geodesic Monte Carlo algorithm \cite{byrne} that has been particularly tailored for sampling orthogonal matrices \cite{tsilifis_rspa}. Such an approach to learn linear embeddings is novel in the context of dimensionality reduction for GPs and clearly prevails previously presented techniques \cite{tripathy_bilionis, garnett} in that the full posterior of the orthogonal matrix can be exploited as opposed to a Laplace approximations while it relieves us of the computational burden caused by high-dimensional gradient-based optimization of the likelihood. 

We structure this paper as follows: Section \ref{sec:argp} presents the basic elements of the classic autoregressive Gaussian Process scheme, Sec. \ref{sec:argp_prior} presents the prior setting and predictive distributions and Sec. \ref{sec:argp_training} discusses the model training approach. Section \ref{sec:argp_w} then formulates the ARGP model defined on a linearly embedded subspace using an orthogonal projection, that is trained using the proposed algorithm presented in Sec. \ref{sec:argp_w_train}. Our numerical examples include two toy problems with known embeddings that are learned using observations from three levels of fidelity (Sec. \ref{sec:ex1} \& \ref{sec:ex2}) and a challenging three-dimensional airfoil optimization problem with a 85-dimensional input space where observations are available from a high- and a low-fidelity simulator.

\section{Recursive Multi-fidelity Gaussian Processes}
\label{sec:mfgp}

\subsection{Autoregressive Gaussian Process model}
\label{sec:argp}

We consider the following scenario where a hierarchy of $s$ computer codes is available, say $\{z_i(\bx)\}_{i=1,\dots,s}$, indexed by input vector $\bx \in \calX \subset \R^D$ where $\calX$ is the design space consisting of all feasible inputs. The codes are in order of increasing fidelity from the cheapest one $z_1(\bx)$, to the most accurate one, $z_s(\bx)$. The key assumption as was first stated by Kennedy \& O'Hagan \cite{kennedyohagan} for any two consecutive levels of code $z_t(\cdot)$ and $z_{t-1}(\cdot)$, is that given observation of the low fidelity code $z_{t-1}(\cdot)$ at $\bx$, nothing more can be learnt about $z_t(\bx)$ by observing $z_{t-1}(\bx')$ at any $\bx' \neq \bx$. This translates to the Markov property 
\begin{equation}
\label{eq:markov}
    \mathrm{cov}\left[z_t(\bx), z_{t-1}(\bx')\vert z_{t-1}(\bx)\right] = 0,
\end{equation}
for any $\bx' \neq \bx$. Based on this principle, we write the relation between any two consecutive codes using the autoregressive model 
\begin{eqnarray}
\label{eq:argp}
\left\{ \begin{array}{l} z_t(\bx) = \rho_{t-1}(\bx) z_{t-1}(\bx) + \delta_t(\bx) + \epsilon_t(\bx)\\
    z_{t-1}(\bx) \bot \delta_{t}(\bx), \ z_{t-1}(\bx) \bot \epsilon_t(\bx), \ z_1(\bx) \bot \epsilon_1(\bx) \\
    \delta_t(\bx) \bot \epsilon_t(\bx), 
\end{array} \right., \ \ 1 < t \leq s
\end{eqnarray}
where ``$\bot$" denotes statistical independence and therefore the discrepancy between successive levels is characterized by an independent correction term $\delta_t(\bx)$ and by the scaling coefficient $\rho_{t-1}$ which captures the correlation between the models, as it satisfies 
\begin{eqnarray}
\rho_{t-1}(\bx) = \frac{\textrm{cov}\left[z_t(\bx), z_{t-1}(\bx)\right]}{\textrm{var}\left[z_{t-1}(\bx)\right]}.
\end{eqnarray}
As suggested in \cite{kennedyohagan}, in what follows we assume for simplicity that $\rho_{t}$ is constant, although dependence on $\bx$ has been modeled using regression functions and was shown to be worthwhile \cite{le_gratiet_juq,le_gratiet_doe}. At last $\epsilon_t(\bx)$ accounts for measurement noise in the obserbations at level $t$ that is assumed to be independent of both $z_{t-1}$ and $\delta_t$.

\subsection{Prior and predictive distributions}
\label{sec:argp_prior}

Prior to observing any outputs of the multi-fidelity codes, we assume that $z_1(\bx)$ and $\{\delta_t(\bx)\}_{t=2,\cdots,s}$ are Gaussian Processes 
\begin{eqnarray}
\label{eq:gp_priors}
\left\{ \begin{array}{ll}
z_1(\bx) \sim \calG\calP \left(0, \sigma_1^2 r_1(\bx, \bx')\right), & \\
\delta_t(\bx) \sim \calG\calP\left( 0, \sigma_t^2 r_t(\bx, \bx') \right), & 1 < t \leq s,
\end{array} \right.
\end{eqnarray}
where $\sigma_t^2$, $t\leq 1, \dots, s$ are scaling factors and $r_t(\bx, \bx')$ are covariance kernel functions. For the measurement noise we take 
\begin{equation}
   \epsilon_t(\bx) \sim \calN(0, \sigma_{\epsilon_t}^2), \ \ \bx\in \calX,\  1\leq t\leq s.
\end{equation}
The kernel functions are modeled using the squared exponential kernel 
\begin{equation} 
r_t(\bx, \bx') = \exp\left[ - \sum_{i=1}^D\frac{ (x_i - x_i')^2}{\theta^2_{i,t}}\right], 
\end{equation}
where $\btheta_t = (\theta_{1,t}, \dots, \theta_{D, t})^T$ are the lengthscales along the $D$ dimensions. The choice of the covariance kernel depends primarily on the prior belief about the smoothness of the response surface (Ch.4, \cite{rasmussen}). 

Let now $D_t = \{\bx_1^{(t)}, \dots, \bx_{n_t}^{(t)}\}$ be the experimental design set at level $t$ that consists of $n_t$ input points in $\calX$, $t=1,\dots, s$ and $\calZ_t := \calZ_t(D_t) = (z_t(\bx_1^{(t)}), \dots, z_t(\bx_{n_t}^{(t)}))^T$ be the observations of the $t$-th code $z_t$. We denote with $\bphi_t$ the parameters introduced at level $t$, so that $\bphi_1 := (\phi_1^1, \phi_1^2, \phi_1^3) = (\btheta_1, \sigma_1^2, \sigma_{\epsilon_1}^2)$ and $\bphi_t := (\phi_t^1, \dots, \phi_t^4) = (\btheta_t, \rho_{t-1}, \sigma_t^2, \sigma_{\epsilon_t}^2)$, $1< t \leq s$ and $\bphi = (\bphi_1, \dots, \bphi_s)$. For a fixed set of parameter values $\bphi$ by stacking all observations to form a vector $\calZ = \left(\calZ_t^T, \dots, \calZ_s^T\right)^T$, we write the predictive distribution of the highest level of code $z_t(D^*)$ at any new set of test points $D^* = \{\bx_i^*\}_{i=1}^{n^*}$, for $\bx^*_i \in \calX$, $i=1, \dots, n^*$ as
\begin{equation}
\label{eq:predictive_GP}
    z_s(D^*)\big\vert \calZ, \bphi \sim \calG\calP(\bm(D^*), \sigma_{Z_s}(D^*))
\end{equation} 
where the predictive mean is given by 
\begin{equation}
\label{eq:m_pred}
    \bm_{Z_s}(D^*) = t_s(D^*) V_s^{-1} \calZ
\end{equation}
and the predictive variance is 
\begin{equation}
\label{eq:s_pred}
    \sigma_{Z_s}^2(D^*) = \sigma^2_{s}(D^*) - t_s(D^*) V_s^{-1} t_s(D^*),
\end{equation}
In the above expressions we have 
\begin{eqnarray}
V_s = \left[\begin{array}{ccc} V^{(1,1)}
     &  \cdots & V^{(1, s)}\\ \vdots 
     & \ddots & \vdots \\ V^{(s, 1)}
     & \cdots & V^{(s, s)} 
\end{array} \right]
\end{eqnarray}
where the diagonal block matrices are given by 
\begin{equation}
    V^{(t, t)} = \sigma_t^2 \left(R_t(D_t)+\sigma_{\epsilon_t}I \right) + \sigma^2_{t-1} \rho_{t-1}^2 \left(R_{t-1}(D_t) +\sigma^2_{\epsilon_{t-1}I} \right) + \dots + \sigma_1^2 \left(\prod_{i=1}^{t-1} \rho_i^2\right) \left(R_1(D_t)+\sigma^2_{\epsilon_1}I\right),
\end{equation}
where $R_i(D_t)$, $i=1, \dots, t$ is the correlation matrix with entries $r_i(\bx, \bx')$, $\bx, \bx' \in D_t$. The off-diagonal blocks are written 
\begin{equation}
    V^{(t, t')} = \left(\prod_{i=t}^{t'-1} \rho_i \right) V^{(t, t)}(D_t, D_{t'}), \ \ 1\leq t<t' \leq s,
\end{equation}
with $V^{(t, t)}(D_t, D_{t'}) = \sigma_t^2 R_t(D_t, D_{t'}) + \dots + \sigma_1^2\left(\prod_{i=1}^{t-1} \rho_i^2\right) R_1(D_t,D_{t'})$. Analogously, $R_i(D_t, D_{t'})$ is the correlation matrix with entries $r_i(\bx, \bx')$, $\bx \in D_t, \bx' \in D_{t'}$. Further, the vector $t_s(D^*)$ is defined as $t_s(D^*) = (\bar{t}_1(\bx, D_1)^T, \dots, \bar{t}_s(\bx, D_s)^T)^T$, where 
\begin{equation}
    \bar{t}_t(D^*, D_t)^T = \rho_{t-1}\bar{t}_{t-1}(D^*, D_t)^T + \left(\prod_{i=t}^{s-1} \rho_{i}\right) \sigma_t^2 R_t(D^*, D_t), \ \ 1 < t \leq s,
\end{equation}
and $\prod_{i=s}^{s-1}\rho_i = 1$ and $\bar{t}_1(D^*, D_1)^T = \left(\prod_{i=1}^{s-1}\rho_i\right) \sigma_1^2 R_1(D^*, D_1)$. At last, the variance $\sigma_{Z_s}^2$ is defined as 
\begin{equation}
    \sigma_{s}^{2}(D^*) = \sum_{t=1}^s \sigma_{t}^2 \left(\prod_{j=t}^{s-1} \rho_j^2\right) r_t(\bx^*, \bx^{*'})
\end{equation}
for $\bx^*, \bx^{*'} \in D^*$.

\subsection{Estimating the model parameters}
\label{sec:argp_training}

In order for the above predictive distribution to be of practical use, it is crucial to train the model by means of finding the optimal set of parameters $\bphi$. Pursuing a fully Bayesian approach to model training, although robust, it can be computationally challenging as the number of levels, and consequently the dimension of $\bphi$, increases. It can be therefore preferable in such cases to resort to more efficient strategies, such as maximum likelihood estimation (MLE). For the sake of completeness, we present below both approaches, that we use interchangeably in our numerical examples. 

\subsubsection{Maximum-likelihood estimation}
\label{sec:mle}

Here we explore the possibility of training the model using maximum likelihood estimation (MLE). More specifically we seek to minimize the negative log-likelihood, that is to identify $\bphi^*$ such that 
\begin{equation}
    \bphi^* =  \arg \min_{\bphi} \bbell(\bphi),
\end{equation}
where 
\begin{equation}
\label{eq:log_like}
    \bbell(\bphi) := - \log p(\calZ\vert \bphi) = \frac{1}{2}\calZ^T V_{s}(\bphi)^{-1}\calZ + \frac{1}{2}\log \vert V_s(\bphi)\vert + \frac{N}{2}\log(2\pi),
\end{equation} 
writing $V_s(\bphi)$ to emphasize the dependence of the covariance matrix on the parameters. Minimization of $\bbell$ can be performed using standard gradient based algorithms. The gradient of $\bbell$ with respect to any of its arguments is given by 
\begin{equation}
\label{eq:grad_like}
    \frac{\partial \bbell(\bphi)}{\partial \phi_t^i} = - \frac{1}{2}\textrm{tr}\left[\left\{ V_s^{-1} \calZ \left(V_s^{-1}\calZ\right)^T - V_s^{-1}\right\} \frac{\partial V_s}{\partial \phi_t^i}\right].
\end{equation}
Note that a convenient simplification applies in the special case considered in \cite{kennedyohagan, le_gratiet_juq, le_gratiet_doe}, where the design points corresponding to the observations are nested, that is $D_t \subset D_{t-1}$. By expanding the likelihood using conditional probabilities and making use of the Markov property in eq. (\ref{eq:markov}) we write 
\begin{equation}
    p(\calZ \vert \bphi) = p(\calZ_s \vert \calZ_{s-1}, \bphi_s) p(\calZ_{s-1}\vert \calZ_{s-2}, \bphi_{s-1}) \cdots p(\calZ_1 \vert \bphi_1), 
\end{equation}
thus the autoregressive model can be trained by solving $s$ distinct optimization problems with respect to the parameters $\bphi_t$, corresponding to the different levels of fidelity $t = 1, \dots s$. At an arbitrary level $t$, the log-likelihood $\bbell_t(\bphi_t) := -\log p(\calZ_t \vert \calZ_{t-1}, \bphi_t)$ is written 
\begin{equation}
    \bbell_t(\bphi_t) = \frac{1}{2} \left(\calZ_t - \rho_{t-1}\calZ_{t-1}(D_t)\right)^T V_s^{-1}\left(\calZ_t - \rho_{t-1}\calZ_{t-1}(D_t)\right) + \frac{1}{2}\log \vert V_s\vert + \frac{n_t}{2}\log (2\pi).
\end{equation}
After differentiating with respect to the components of $\bphi_t$ and setting equal to zero, one can derive the maximum-likelihood estimates
\begin{equation}
    \hat{\rho}_{t-1} = \left[\bh_t^T (R_t(D_t) + \sigma_{\epsilon_t}^2)^{-1} \bh_t\right]^T \bh_t^T \left(R_t(D_t) + \sigma_{\epsilon_t}^2\right)^{-1} \calZ_t,
\end{equation}
and 
\begin{equation}
    \hat{\sigma}_t^2 = \frac{1}{c} \left(\calZ_t - \hat{\rho}_{t-1}\calZ_{t-1}(D_t)\right)^T \left(R_t(D_t) + \sigma_{\epsilon_t}^2\right)^{-1}\left(\calZ_t - \hat{\rho}_{t-1}\calZ_{t-1}(D_t)\right),
\end{equation}
where $\bh_t = \left[\mathbf{1}_{n_t}\ \calZ_{t-1}(D_t) \right]$, $c = (n_t -1)\mathbb{I}_{t=1} + (n_t - 2)\mathbb{I}_{t>1}$, $\calZ_{t-1}(D_t)$ is the set of observations from code $z_{t-1}$ corresponding only to the design points in $D_t$. Further, $\mathbf{1}_{n_t}$ is a vector of length $n_t$, filled with ones and $\mathbb{I}_A$ is the indicator function that is one in $A$, and zero otherwise. At last, the two estimates given above are dependent on the lengthscales $\btheta_t$ and the noise variance $\sigma_{\epsilon_t}^2$. Those can be estimated by maximizing the concentrated restricted log-likelihood function 
\begin{equation}
\bbell^{res}_{t} := \log \vert R_t(D_t) + \sigma_{\epsilon_t}^2\bI \vert + c\log \hat{\sigma}^2_{t},\ \  t = 1, \dots, s. 
\end{equation}

\subsubsection{Markov Chain Monte Carlo sampling}
\label{sec:mcmc}

A fully Bayesian updating strategy of the model parameters can be carried out using Markov Chain Monte Carlo (MCMC) sampling in order to generate samples of the posterior distribution of $\bphi$. The latter is written as 
\begin{equation}
    p(\bphi \vert \calZ) \propto p(\calZ \vert \bphi)p(\bphi)
\end{equation}
where the likelihood distribution is $p(\calZ \vert \bphi) = \exp(-\bbell(\bphi))$ with $\bell(\bphi)$ given in eq. (\ref{eq:log_like}). The prior distributions of parameters $\phi$ corresponding to lengthscales at various levels of fidelity are modeled using independent Beta distributions and those corresponding to the variances are modeled using independent inverse gamma distributions. In our implementations we use a Metropolis-Hastings algorithm \cite{gelman} that allows jumps in order to fully explore possible multimodal behavior.

\section{Multi-fidelity Gaussian Processes on low-dimensional embeddings}
\label{sec:mfgp_w}

\subsection{Dimensionality reduction using projection matrices}
\label{sec:argp_w}

As discussed in the introduction, the main goal in this paper is to learn a response surface on a high dimensional input space within a multi-fidelity context, i.e. by leveraging observations from low accuracy simulators that are cheaper to evaluate. Several shortcomings can make the learning process problematic in the presence of high dimensions and large datasets. For instance, the large number of parameters in the case of anisotropic kernels can result in poor performance of the MCMC algorithm or convergence of the MLE procedure to suboptimal solutions. Below, we develop an dimensionality reduction framework where the ARGP model is trained on a low dimensional input space that is the result of a linear embedding of the original space. We assume throughout this work that the target function $z_s :\R^D \in \R$ can be described or be well-approximated by a function $f_s : \R^d \to \R$, defined in a d-dimensional space $\calX_d$, where $d \ll D$, such that
\begin{equation}
 z_s(\bx) \approx f_s(\bW^T \bx). 
\end{equation}
Here, $\bW$ is assumed to be a $D \times d$ orthonormal matrix that maps the original design space $\calX$ to $\calX_d$. The choice of orthonormality is made so that the columns of $\bW$ form a basis on $\calX_d$ and therefore the matrix itself is a projection from $\calX$ to $\calX_d$. That further implies that once $\calX_d$ is identified, any other set of basis vectors forms a projection that can describe the same approximation of $z_s$. 

Next, it is important to assume that all the lower fidelity codes can be approximated by similar ``link" functions defined on the same low dimensional space $\calX_d$, that is using the same projection function $\bW$. We therefore assume that for each $t =1, \dots, s-1$
\begin{equation}
    z_t(\bx) \approx f_t(\bW^T \bx),
\end{equation}
for functions $f_t :\R^d \to \R$, $t=1, \dots, s-1$. Thus, we can define the autoregressive model (\ref{eq:argp}) on $\calX_d$ as 
\begin{eqnarray}
\label{eq:argp_w}
\left\{ \begin{array}{l} f_t(\Tilde{\bx}) = \rho_{t-1}(\tilde{\bx}) f_{t-1}(\tilde{\bx}) + \delta_t(\tilde{\bx}) + \epsilon_t(\tilde{\bx})\\
    f_{t-1}(\tilde{\bx}) \bot \delta_{t}(\tilde{\bx}), \ f_{t-1}(\tilde{\bx}) \bot \epsilon_t(\tilde{\bx}), \ f_1(\tilde{\bx}) \bot \epsilon_1(\tilde{\bx}) \\
    \delta_t(\tilde{\bx}) \bot \epsilon_t(\tilde{\bx}), 
\end{array} \right., \ \ 1 < t \leq s,
\end{eqnarray}
where $\tilde{\bx}= \bW^T\bx$ and $\bx \in \calX$.

The idea behind this formulation is that, by identifying $\bW$ such that the above approximations are accurate, the multi-fidelity output quantify of interest is described as a function of a low dimensional input, thus, it becomes simpler to characterize its predictive distribution. Applying the same projection $\bW$ at all levels of fidelity, practically means that all codes exhibit most of their variability within the same ``active" subspace, as it was termed by Constantine \cite{constantine_AS, constantine_book}. Although this might seem as a strong assumption, in fact, considering that different fidelity codes are typically highly correlated as they simulate the same physical process at different levels of accuracy, the assumption is fairly plausible. Furthermore, it is worth pointing out that the ARGP model (\ref{eq:argp_w}) where the low dimensional spaces are defined using different $\bW_t$, $t = 1, \dots, s$ at different levels, does no longer honor the Markov property (\ref{eq:markov}). We do not pursue further such a scenario in this work. 

Assigning the same prior distributions as in eq. (\ref{eq:gp_priors}) for the reduced dimensionality ARGP introduced above, results in the same posterior expressions given in eqs. (\ref{eq:m_pred})-(\ref{eq:s_pred}), where the covariance matrices describe the correlations of the training points in $\calX_d$. By denoting the projections of all design points where the model outputs are observed, as $D^{\bW}_t = \{\tilde{\bx} = \bW^T \bx: \bx \in D_t\}$, $t = 1, \dots s$, we can rewrite the covariance kernels $R_i(D_t^{\bW})$ and $R_i(D_t^{\bW}, D_{t'}^{\bW})$ as functions of the high dimensional inputs in the original space $\calX$, with entries
\begin{eqnarray}
    \begin{array}{ccc} r_i(\tilde{\bx}, \tilde{\bx}') = r_i(\bW^T\bx, \bW^T\bx'), \ \ \bx, \bx'\in D_t, & \textrm{and} & r_i(\tilde{\bx}, \tilde{\bx}) = r_i(\bW^T\bx, \bW^T\bx'), \ \ \bx\in D_t, \bx'\in D_{t'},
    \end{array}
\end{eqnarray}
respectively. By incorporating these expressions in the likelihood and posterior distributions, it becomes clear that $\bW$ can be considered as an additional set of model parameters that needs to be inferred from observations, in a similar manner as in the single-fidelity setting presented in \cite{tripathy_bilionis}. 

\subsection{Simultaneous autoregressive GP training and embedding learning}
\label{sec:argp_w_train}

As highlighted above, training the autoregressive GP model and identifying the low-dimensional design space requires learning the model parameters $\bphi$ and the rotation matrix $\bW$ simultaneously. To do so, we propose a two-step iterative procedure that iterates between tuning the $\bphi$ parameters for a fixed $\bW$ and updating $\bW$ while keeping the parameters $\bphi$ fixed. Such algorithms have been used in the past for dimensionality reduction purposes within the context of GP regression \cite{tripathy_bilionis, garnett} and Polynomial Chaos adaptations \cite{tsilifis_rspa, tsilifis_cs} and have demonstrated great potential. To further justify the choice of updating scheme, we can write the Bayesian posterior of the joint parameters $(\bphi, \bW)$ given observations $\calZ$, as 
\begin{equation}
p(\bphi, \bW \vert \calZ) \propto p(\calZ \vert \bphi, \bW)p(\bphi)p(\bW).
\end{equation}
Assuming that sampling from the marginal posteriors of $\bphi$ and $\bW$ conditional on each other, a Gibbs sampler would consist of generating a Markov chain that eventually converges to the joint posterior above \cite{chaspari}, therefore generating a chain of $\bphi_n$ and $\bW_n$ samples in such a fashion will ultimately explore $p(\bphi, \bW \vert \calZ)$.

In order to carry out such a sampling scheme, we employ the following strategy: When $\bW$ is given, the $\bphi$ updating step is performed using the methods presented in Section \ref{sec:argp_training}. For updating $\bW$ while $\bphi$ is kept fixed, we use the Geodesic Monte Carlo method that samples from target distributions defined on embedded manifolds, as it is described in Byrne \& Girolami \cite{byrne}. In our case, $\bW$ is a matrix that takes values on the Stiefel manifold of orthonormal $d$-frames on $\R^D$ \cite{muirhead}, and the target density is the marginal posterior $p(\bW \vert \bphi, \calZ)$. The iterative procedure is summarized in Algorithm \ref{alg:two_step}. The details of Geodesic Monte Carlo sampling are presented in the next section. 

\begin{algorithm}[h]
\caption{Two-step iterative update of $\bphi$ and $\bW$ \label{alg:two_step}}

\SetKwInOut{Input}{Input}\SetKwInOut{Require}{Require}
\DontPrintSemicolon
\Require{Design input sets $\{D_t\}_{t=1, \dots, s}$, observations $\{\calZ_t\}_{t=1, \dots, s}$, initial guess $\bW_0\sim p(\bW)$, assign priors on $\bphi$ or initialize to $\bphi_0$.}
\Repeat{relative change in Hamiltonian function (\ref{eq:hamiltonian}) is less than tolerance $\epsilon_H$. }{
$\bphi^{(n)} \leftarrow$ Run MCMC or MLE optimization (Sec. \ref{sec:mcmc}-\ref{sec:mle}) \\
$\bW^{(n)} \leftarrow$ Run Geodesic MC Algorithm \ref{alg:geod_mc} with target density $p(\bW\vert \calZ, \bphi^{(n)}) \propto p(\calZ\vert \bW, \bphi^{(n})p(\bW)$
}
\end{algorithm}

\subsubsection{Geodesic Monte Carlo}

The Geodesic Monte Carlo algorithm developed in \cite{byrne} is at its core a Hamiltonian Monte Carlo (HMC) sampling technique \cite{girolami_hmc} defined on a Riemannian manifold embedded in $\R^D$. A Hamiltonian function is defined that describes the dynamics of a spatial variable and is characterized by its target density. Next, the Hamiltonian flows are simulated through numerical integration in order to propose new samples that are to be accepted or rejected, based on a Metropolis-Hastings step \cite{hitchcock}. In our case, we are working on the Stiefel manifold that is defined as 
\begin{equation}
    \calV_{d, D} = \{\bw \in \R^{D\times d}: \bw^T\bw = \bI_d\},
\end{equation}
for $d\leq D$, where the special cases $d = D$ and $d=1$ correspond to the set of all orthonormal square matrices and the ($D-1$)-dimensional hypersphere on $\R^D$, respectively. The Hamiltonian corresponding to our target posterior density is defined as 
\begin{equation}
\label{eq:hamiltonian}
    H(\bw, \bu) := H^{[1]}(\bw, \bu) + H^{[2]}(\bw, \bu) = - \log p(\bw \vert \bphi, \calZ) + \frac{1}{2}\bu^T\bu,
\end{equation}
where $\bu \in \R^{D\times d}$ is an auxiliary velocity variable and the dynamics of $H(\bw, \bu)$ are described by
\begin{eqnarray}\left\{
\begin{array}{crl}
    \dot{\bw} = & \displaystyle{\frac{\partial H}{\partial \bu}} = &  \bu\\
    \dot{\bu} = & - \displaystyle{\frac{\partial H}{\partial \bw}} = & \nabla_{\bw} \log p(\bw \vert \bphi, \calZ)
\end{array} \right. .
\end{eqnarray}
The key difference of the sampling strategy proposed in \cite{byrne} from traditional HMC algorithms is that instead of relying on symplectic integrators \cite{neal} to simulate the Hamiltonian flow, we take advantage of the fact that the dynamics of the kinetic term $H^{[2]}(\bw, \bu)$ describe a flow over a geodesic curve that are explicitly known for certain manifolds, therefore numerical integration applies only on the potential term, thus improving accuracy and performance.

In summary, at the $n$-th step of the algorithm, a random velocity vector is proposed that is tangent on the manifold at the previously accepted step $\bW^{(n-1)}$, and it specifies the direction along which the Hamiltonian is going to move. Then the two terms $H^{[1]}(\bw, \bu)$ and $H^{[2]}(\bw, \bu)$ are integrated over time $t=\epsilon$ by first updating $H^{[1]}(\bw, \bu)$ for a time step $t=\epsilon/2$, followed by updating $H^{[2]}(\bw,\bu)$ for $t=\epsilon$ using the known geodesic curve formula, and then $H^{[1]}(\bw, \bu)$ is integrated again for $t=\epsilon/2$. The procedure is repeated until a user-defined final time $t=T$ is reached. At last, the resulting ``spatial" coordinate $\bw^*$ will be accepted in the chain with probability
\begin{equation}
    \alpha_{acc} = \min \left\{1, \exp\left[- H(\bw^*, \bu^*) + H(\bw_0, \bu_0)\right]\right\}.
\end{equation} 
The full mathematical details of the integrator scheme for the Hamiltonian flow and the geodesic formulas on $\calV_{d, D}$ are given in \ref{sec:geod_mc_math}. Algorithm \ref{alg:geod_mc} describes one complete accept-reject step of the above procedure.
As a prior on $\bW$, we use a Matrix-Langevin (mL) distribution \cite{chikuse} whose density function is given by 
\begin{equation}
    p(\bw) = \frac{1}{c(\bF)}\exp\left\{ \textrm{Tr}\left[\bF^T \bw\right] \right\},
\end{equation}
where $c(\bF)$ is the normalizing constant that is parametrized by the matrix $\bF \in \R^{D\times d}$. Details on the geometric interpretation of the mL density and how to tune the prior parameters $\bF$ are given in \ref{sec:mL}. Note that updating $\bu$ by integrating the potential term $H^{[1]}(\bw, \bu)$ requires computing the gradient of the marginal log-posterior distribution $p(\bW \vert \bphi, \calZ) \propto p(\calZ \vert \bphi, \bW) p(\bW)$.
The gradient of the marginal log-posterior becomes
\begin{equation}
    \nabla_{\bw} \log p(\bW \vert \bphi, \calZ) = \nabla_{\bw} \log p(\calZ \vert \bphi, \bW) + \nabla_{\bw} \log p(\bW) 
\end{equation}
where the likelihood gradient is given by (\ref{eq:grad_like}) and the gradient of $V_s$ consists of gradients of block matrices  $\frac{\partial V^{t, t'}}{\partial w_{ij}}$, $1\leq t \leq t' \leq s$. Those involve differentiating $R_i(D^{\bW}_t, D^{\bW}_t)$ that have $(i,j)$-th entries
\begin{equation}
    \frac{\partial r_i(\tilde{\bx}, \tilde{\bx}')}{\partial w_{ij}} = \frac{\partial}{\partial \tilde{\bx}_j}\left[ r_i\left(\bW^T\bx, \bW^T\bx'\right) \right]\bx_i + \frac{\partial}{\partial \tilde{\bx}'_j}\left[r_i\left(\bW^T\bx, \bW^T\bx'\right)\right] \bx'_i.
\end{equation}
At last, the log-prior gradient is $\nabla_{\bw}\log p(\bW) = \bF^T$.  

\begin{algorithm}[h]
\caption{Geodesic Monte Carlo algorithm \cite{byrne}\label{alg:geod_mc}}
\SetKwInOut{Initialize}{Initialize}
\Initialize{Choose integration period $T$, time step $\epsilon$ and sample $\bW_0 \sim p(\bW)$. \\
At the $n$-th step assume $\bW_n = \bW$:}
$\bu \sim \calN(0, \bI_{D, d})$ \\
$\bu \leftarrow \Pi_{\bW}(\bu)$ \\
$H \leftarrow \log p(\bW | \bphi, \calZ) - \frac{1}{2}\bu^T\bu$ \\
$\bW^* \leftarrow \bW$\\
\For{$h = 1$ to $T$}{
	$\bu \leftarrow \bu + \frac{\epsilon}{2}\nabla_{\bW}\log p(\bW^* | \bphi, \calZ)$\\
	$\bu \leftarrow \Pi_{\bW}(\bu)$\\
	Update $(\bW^*, \bu)$ by following the geodesic flows (\ref{eq:geod_nd})-(\ref{eq:geod_1d}) for a time interval $\epsilon$\\
	$\bu \leftarrow \bu + \frac{\epsilon}{2}\nabla_{\bW}\log p(\bW^* | \bphi, \calZ)$\\
	$\bu \leftarrow \Pi_{\bW}(\bu)$ 
}
$H^* \leftarrow \log p(\bW^*\vert \bphi, \calZ) - \frac{1}{2}\bu^T\bu$\\ 
$u \sim \calU(0,1)$\\
\If{ $u < \exp\left(H^* - H\right)$}{
$\bW \leftarrow \bW^*$
}
\end{algorithm}

\section{Numerical examples}
\label{sec:numerics}

For all numerical examples presented in this section we have used the following settings: 
\begin{enumerate}
    \item The number of time steps and the step-size used in the integration of the Hamiltonian in Algorithm \ref{alg:geod_mc} are set to $T = 10$ and $\epsilon = 0.05$ respectively. These values have been carefully selected after numerical experimentation with the algorithm and they provide a moderate acceptance rate while keeping running times to reasonably low levels. Intuitively, a highly accurate integration resulting in high acceptance rate would require a large value for $T$ and very small $\epsilon$ but it would slow down the algorithm significantly. On the other hand, coarse integration using large $\epsilon$ and small $T$ would speed up the algorithm but would reduce the acceptance rate. Another characteristic of the algorithm that manifests in high dimensions for both $d$ and $D$ is that for an arbitrary initial point $\bW_0$ the acceptance rate during the first few iterations is significantly low, as a result of the standard Gaussian proposal from which we sample $\bu$, leading to completely random directions along with the Hamiltonian flow evolves. To improve performance of the algorithm, we change the mL prior at each iteration $n$ by setting its parameter $\bU = \bW_{n-1}$. Each iteration stops when one single sample $\bW$ is accepted and we reduce the step size $\epsilon$ to $\epsilon / 1.2$ after $20$ successive rejections in order to refine the Hamiltonian integration and reset on the next iteration. 
    We stop sampling as soon as the first sample is accepted.
    \
    \item To train the ARGP using MCMC sampling, we assign beta priors with parameters $(1, 0.1)$ on the lengthscales $\btheta_t$ and inverse Gamma priors with parameters $(5,5)$ and $(1, 10^{-4})$  on the variance parameters $\sigma_t^2$ and $\sigma_{\epsilon_t}^2$ respectively, for $t = 1, \dots, s$. To proceed with sampling $\bW$, the hyperparameters are then fixed to the median values of the full MCMC chain after 200 samples are accepted. The MCMC algorithm is run using General Electric Global Research Center's in-house Bayesian modeling toolbox GEBHM \cite{ghosh2020advances}. To train the ARGP using MLE, the log-likelihood or the restricted log-likelihoods are maximized using the BFGS algorithm \cite{byrd}.
\end{enumerate}

\subsection{Academic example 1: Three-fidelity model with known 1-dimensional embedding}
\label{sec:ex1}

We consider the following three levels of code 
\begin{eqnarray}
\label{eq:ex1_codes}
\left\{
\begin{array}{rl}z_1(\bx) = & f_1(\bw^T \bx) \\
z_2(\bx) = & f_2(\bw^T\bx) \\ 
z_3(\bx) = & f_3(\bw^T\bx)
\end{array} \right. ,
\end{eqnarray}
where $\bw \in \calV_{1, D}$ so that $\bw^T\bx$ is a scalar variable and the \emph{link} functions $f_1, f_2, f_3$ are given by
\begin{eqnarray}
\left\{
\begin{array}{l}
     f_1(\bw^T\bx) = \frac{1}{2}(8\bw^T\bx - 2)^2 \sin(5\bw^T\bx - 4) + 10(\bw^T\bx - 1/2)\\
     f_2(\bw^T\bx) = 2 f_1(\bw^T \bx) - 20 \bw^T\bx + 20 \\
     f_3(\bw^T\bx) = \frac{3}{2} f_2(\bw^T\bx) + 30 (\bw^T\bx)^2
\end{array} \right. .
\end{eqnarray}

For this example we take $D = 10$ and we generate $\bw$ randomly, by fixing the random seed in order to ensure reproducibility. 
For our numerical experiments, the projection matrix is fixed to 
\begin{eqnarray}
\bw = \left[\begin{array}{r}
     0.14042\\ -0.35474\\ 0.42674\\ -0.09312\\ -0.21463\\ 0.26425\\
  0.25603 \\ -0.18959\\ 0.00467\\ -0.66800
\end{array} \right].
\end{eqnarray}

\begin{figure}[h]
    \centering
    \includegraphics[width=0.65\textwidth]{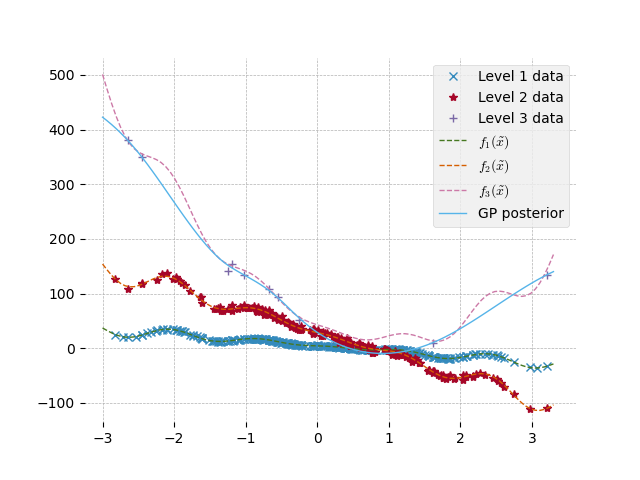}
    \caption{Academic example 1. Training data along with the 1-dimensional representation of $z_1$, $z_2$ and $z_3$ and the GP posterior mean mean using only $D_3$ data.}
    \label{fig:ex1_data}
\end{figure}

We generate synthetic data that consists of observations obtained on nested design points $D_1, D_2$ and $D_3$ from the low, intermediate and high fidelity codes respectively, where $\vert D_1 \vert=300$, $\vert D_2\vert = 200$ and $\vert D_3\vert = 10$. The observations are corrupted by Gaussian noise at all levels of fidelity with their standard deviations being equal to $0.5$, $3$ and $5$, corresponding to $z_1$, $z_2$ and $z_3$ respectively. Fig. \ref{fig:ex1_data} shows the full training data set along with the true 1-dimensional link functions $f_1$, $f_2$ and $f_3$. To motivate our study, a single-level Gaussian process regression is performed on the $10$ high-fidelity data points that are available. The predictive mean, depicted with solid blue line, matches the training points but clearly is unable to capture the true model's fluctuations due to the absence of a sufficient amount of data, let alone the fact that the true $\bw$ and the resulting 1-dimensional representation is hypothetically not available to the experimenter. That essentially means that a dimensionality reduction approach performed using only the high fidelity dataset, as proposed in \cite{tripathy_bilionis} would provide such a predictive mean only in the best case scenario, where $\bw$ would be recovered exactly. In addition. even learning this rotation matrix would be challenging due to the limited data availability and convergence of the two step algorithm would be expected to be extremely slow. Lastly, it is needless to say that gradient-based methods such as \cite{constantine_AS} would simply fail dramatically due to the poor gradient Monte Carlo estimate using only $10$ data points. On the other hand, in a multi-fidelity GP regression setting in the original 10-dimensional design space, the available data might fail to provide meaningful inference results, while training the model becomes again challenging due to the large number of parameters to be inferred, including different lengthscales that are present in the anisotropic kernels, as well as the repeated use of Cholesky decomposition for inverting large covariance matrices. 

\begin{figure}[ht]
    \centering
    \includegraphics[width=0.49\textwidth]{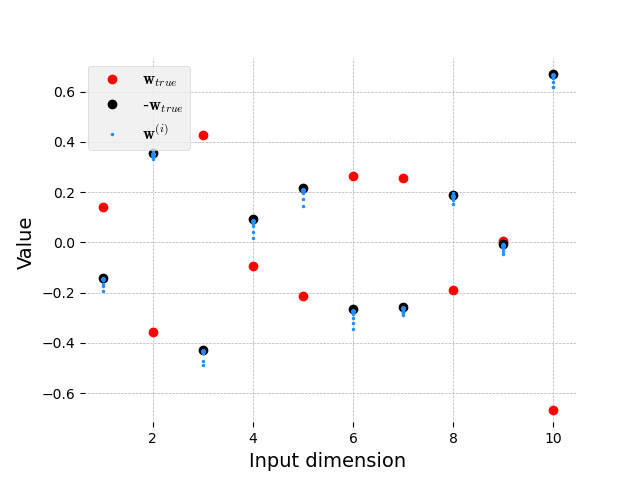}
    \includegraphics[width=0.49\textwidth]{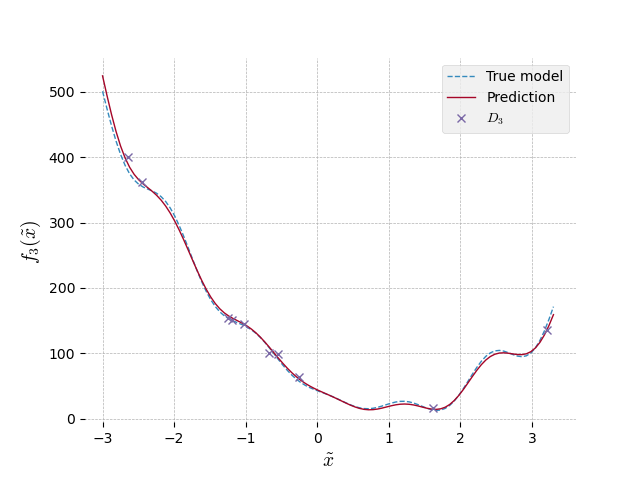}
    \caption{Academic example 1. Left: Entry values of the posterior samples $\bw^{(i)}$ and the true values of $\bw$ and $-\bw$. Right: Comparison of the multi-fidelity predictive mean and the true link function $f_3(\tilde{x})$. }
    \label{fig:ex1_w}
\end{figure}

We run Algorithm \ref{alg:two_step} for this particular setting and after only eleven iterations we obtain the converged rotation. Fig. \ref{fig:ex1_w} (left) shows the drawn samples during iterations 5-11. It can be seen that the values are in full agreement with -$\bw$ which is a valid rotation since the representation of $z_3$ through its link function $f_3$ is invariant under reflections about the origin. Fig. \ref{fig:ex1_w} (right) shows the plots of the predictive mean $f_3(\tilde{x})\vert \calZ$ and the true link function $f_3(\tilde{x})$ from eq. $(\ref{eq:ex1_codes})$. The excellent agreement between the two is apparent. At last, Table \ref{tab:ex1_mle} shows the estimated model parameter values $\bphi$. For comparison we display the value of the single-level GP regression shown in Fig. \ref{fig:ex1_data}. As one can conclude from the plot, the single-level GP overestimates the lengthscale $\hat{\theta_3} = 5.223$ and interprets the data discrepancies as observation noise ($\log\sigma_{\epsilon_3}^2 = 4.017$). On the contrary, the obtained ARGP gives high fidelity lengthscale $\hat{\theta_3}=1.7$ while the noise variances for all fidelities are almost negligible. By capturing accurately the correlations between the different levels $\hat{\rho}_1 = 1.955$ and $\hat{\rho}_2 = 1.242$ (recall the true values $\rho_1 = 2$ and $\rho_2 = 1.5$), the autoregressive model leverages the low fidelity data effectively and captures the model's fluctuations even in areas such as the interval $\tilde{x} \in [2, 3]$ where high fidelity training points are fully absent. 

\begin{table}[h]
    \centering
    \begin{tabular}{c|ccccccccccc}
       & $\theta_1$ & $\log\sigma_1^2$ & $\log \sigma_{\epsilon_1}^2$ & $\rho_1$ & $\theta_2$ & $\log\sigma_2^2$ & $\log\sigma_{\epsilon_2}^2$& $\rho_2$ & $\theta_3$ & $\log\sigma_3^2$ & $\log\sigma_{\epsilon_3}^2$ \\ \hline \hline
        \small GP & & & & & & & & & \small 5.223 & \small 11.601 & \small 4.017 \\ \hline 
        \small ARGP & \small -0.508 & \small 7.022 & \small -8.297 & \small 1.955 & \small 3.296 & \small 11.518 & \small -17.868 & \small 1.242 & \small 1.700 & \small 13.107 &\small -20.123
        
    \end{tabular}
    \caption{Academic example 1. Final maximum likelihood estimates of the model parameters.}
    \label{tab:ex1_mle} 
\end{table}

\subsection{Academic example 2: Three-fidelity model with known 2-dimensional embedding}
\label{sec:ex2}

A three-level autoregressive Gaussian Process is presented in this example where the low dimensional embedding $\calX_d$ is now 2D and the additional challenge of identifying the dimensionality of $\calX_d$ is also explored. Let $\bW = [\bw_1\ \bw_2] \in \calV_{2, D}$ be a fixed projection matrix with columns $\bw_1$ and $\bw_2$. The three levels of code are given by the functions 
\begin{eqnarray}
\left\{
\begin{array}{l}
     f_1(\bW^T\bx) = \sin(\bw_1^T \bx)\\
     f_2(\bW^T\bx) = f_1(\bW^T\bx) - 7\sin^2(\bw_2^T\bx) \\
     f_3(\bW^T\bx) = \frac{3}{2}f_2(\bW^T\bx) + 5 \left(\bw_2^T\bx\right)^2 \sin(\bw_1^T\bx)
\end{array} \right. .
\end{eqnarray}
We consider again $D=10$ to be the dimensionality of $\calX$ and the projection matrix is fixed to
\begin{eqnarray}
\bW = \left[\begin{array}{rr}
 0.28490 &  0.34201 \\
	 -0.21608 &  0.19310 \\
	 -0.46249 &  0.36223 \\ 
	 -0.15187 & -0.05088 \\
	 -0.16601 & 0.51910 \\
	  0.70297 & 0.23900 \\
	 -0.16004 & 0.23084 \\
	  0.06096 & -0.48747\\ 
	  0.23763 & 0.26276 \\ 
	 -0.16620 & -0.15930
\end{array}\right].
\end{eqnarray}
For this example we generate data again from nested design points $D_1$, $D_2$ and $D_3$ where this time $\vert D_1\vert = 200$, $\vert D_2\vert = 100$ and $\vert D_3\vert = 25$. Observations are contaminated with Gaussian noise whose standard deviation is equal to $\sigma_{\epsilon_1} = 0.1$, $\sigma_{\epsilon_2} = 0.1$ and $\sigma_{\epsilon_3} = 0.05$ at the corresponding levels.

\begin{figure}[t]
    \centering
    \includegraphics[width=0.49\textwidth]{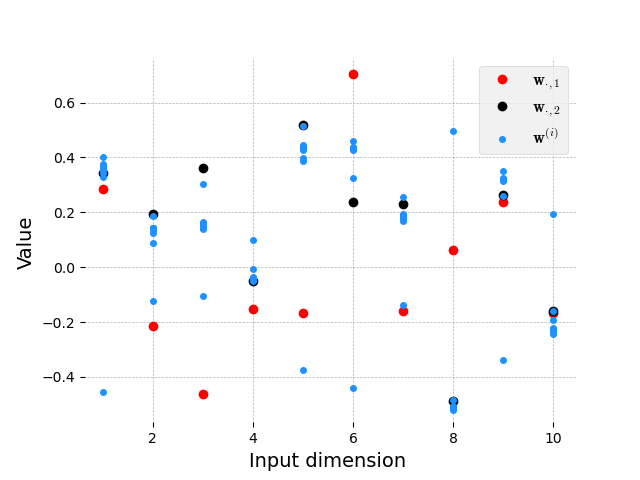}
    \includegraphics[width=0.49\textwidth]{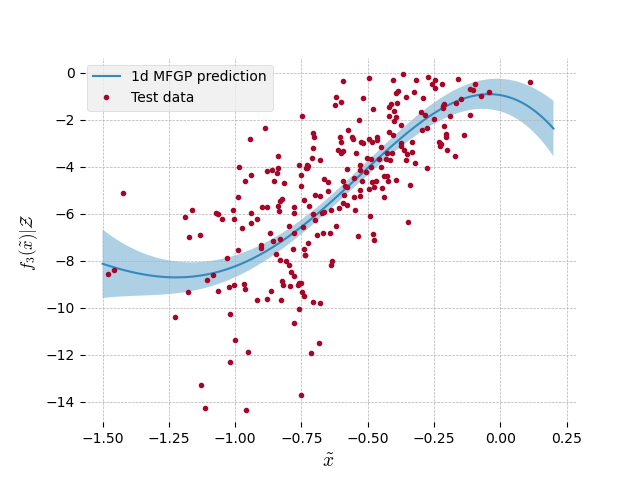}
    \caption{Academic example 2. Left: Comparison of samples $\bw^{(i)}$ versus values of columns of true $\bw$. Right: Predictive mean and 3-standard deviation confidence bands of the 1-dimensional adapted MFGP along with 250 test data projected in the embedded space.}
    \label{fig:ex2_1d}
\end{figure}

\begin{table}[h]
    \centering
    \begin{tabular}{c|ccccccccccc}
       & $\btheta_1$ & $\log\sigma_1^2$ & $\log \sigma_{\epsilon_1}^2$ & $\rho_1$ & $\btheta_2$ & $\log\sigma_2^2$ & $\log\sigma_{\epsilon_2}^2$& $\rho_2$ & $\btheta_3$ & $\log\sigma_3^2$ & $\log\sigma_{\epsilon_3}^2$ \\ \hline \hline
        \small d=1 & \small 2.005 & \small  0.651 & \small -3.082 & \small 2.419 & \small 0.327 & \small 4.901 & \small -6.903 & \small 2.505 & \small 2.510  & \small 8.807 &\small -21.089 \\ \hline
        \small d=2 & \small (2.67, 4.40) & \small 4.932 & \small -15.004 & \small 0.123 & \small (2.96, 0.26) & \small 4.756 & \small -8.888 & \small -0.529 & \small (0.49, -0.31) & \small 4.083 & -18.466

    \end{tabular}
    \caption{Academic example 2. Final maximum likelihood estimates of the model parameters.}
    \label{tab:ex2_mle} 
\end{table}

\begin{figure}[h]
    \centering
    \includegraphics[width=0.49\textwidth]{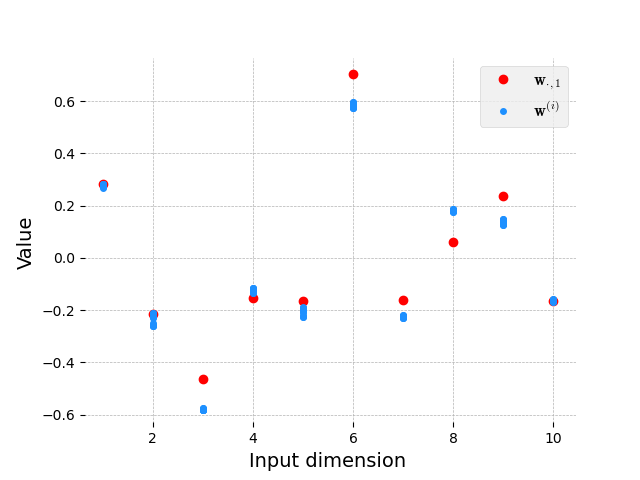}
    \includegraphics[width=0.49\textwidth]{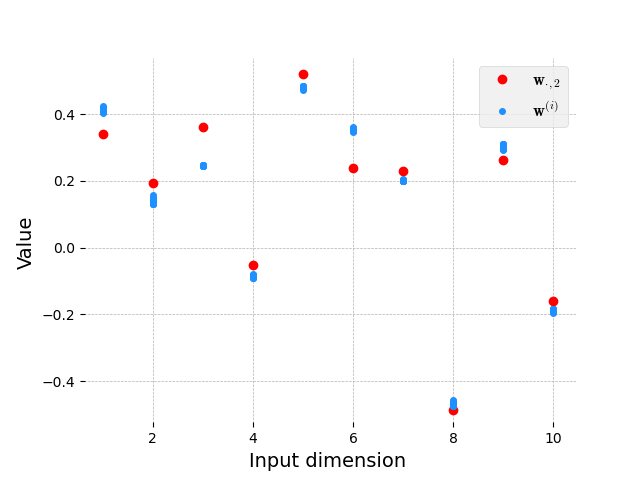}
    \includegraphics[width=0.49\textwidth]{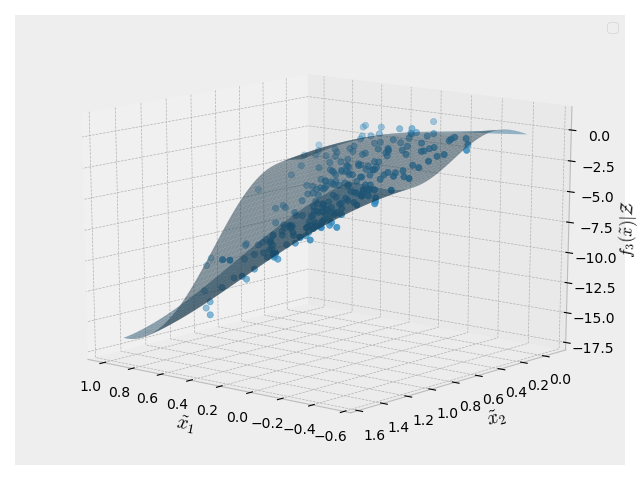}
    \includegraphics[width=0.49\textwidth]{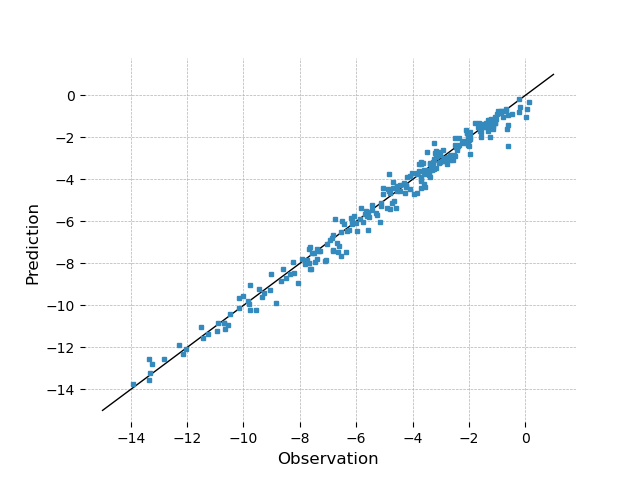}
    \caption{Academic example 2. Top: Comparison of posterior samples of $\bw_1$ and $\bw_2$ columns of $\bW$ versus the nominal values used to generate the training data. Bottom left: Predictive posterior mean of the trained MFGP model as a function of its two arguments $\tilde{x}_1$, $\tilde{x}_2 \in \calX_2$ along with $250$ test observations generated from the true model $z_3(\bx)$. Bottom right: Prediction at the same $250$ test points versus observations. Perfect predictions would fall on the $45^{\circ}$ line depicted in black solid color.}
    \label{fig:ex2_2d}
\end{figure}

We run again Algorithm \ref{alg:two_step}, first for $d=1$ and then for $d=2$ and we report our results below. For the $d=1$ case, the algorithm converged after roughly 12 iterations and, as expected, the trained model fails to capture a suitable embedded space that could honor all training data points and further represent the high fidelity code as a 1-dimensional function. Fig. \ref{fig:ex2_1d} (left), compares the posterior values of $\bW$ accepted while running the Geodesic MC algorithm, with the values of the two columns of the true $\bW$ used to generate the training data, Fig. \ref{fig:ex2_1d} (right) shows the 1-dimensional predictive mean of the posterior ARGP with 3-standard-deviation-wide confidence bands along with $250$ test data points. One can observe that the entries of the inferred $\bw$ tend mostly towards the values of the 2nd column of the true $\bW$. This can be intuitively explained from the fact that the terms including $\bw_2$ appear to be more dominant in the overall expression of $f_3(\bW^T\bx)$ and particularly the low fidelity model $f_1(\bW^T\bx)$ has small impact on the high fidelity code. At the same time, the predictive capabilities of the posterior model clearly fail to span the regions where the additional test data might be observed, indicating that a higher dimensional embedding should be learned. In the $d=2$ case, the situation improves significantly. As can be seen in Fig. \ref{fig:ex2_2d}, top row, the sampled posterior values of $\bw_1$ and $\bw_2$ are in agreement with the true values of $\bW$ in both columns. The samples have been obtained during iterations 10-20. Furthermore, the bottom row graphs show the 2-dimensional predictive mean of the train ARGP model along with $250$ test data points and a 45-degree line plot comparing observations versus predictions on the same data points. The overall predictive performance of the model is in agreement with the true model output.

\section{Airfoil shape optimization problem}
\label{airfoil_problem}

The aerodynamic optimization problem analyzed in this section is the full 3D design of a typical large last stage blade (LSB) of an industrial gas turbine (IGT). As the largest rotating component in the turbine, last stage blades are one of the most mechanically challenged components in an IGT and often determine the total power output of the machine. Increasing push for larger and hotter turbines to drive down cost of electricity has resulted in increasingly challenging LSB designs. With deference typically skewed towards durability, this has generally resulted in greater aerodynamic compromises that negatively impact blade efficiency. This also drives longer design cycles as designers incrementally search for acceptable aero-mechanical solutions. In this type of design, aerodynamic assessments of power, efficiency, and flow capacity are based on expensive high-fidelity computational fluid dynamics (CFD) simulations. Each design iteration requires one or more CFD simulations, depending on the number of inner aerodynamic iterations required to satisfy the cycle's flow capacity requirements. A full 3D optimization would enable the enhancement of turbine performance. However, the process normally implies a large computational cost because of the high dimensionality of the design space.

\begin{figure}[h]
    \centering
    \includegraphics[width=0.8\textwidth]{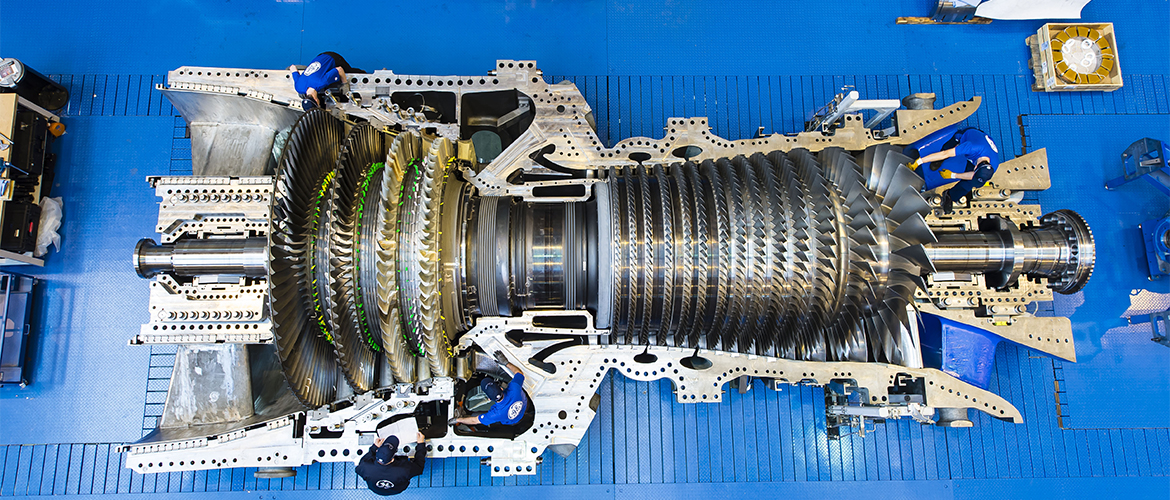}
    \caption{9HA Industrial Gas Turbine (\url{https://www.ge.com/power/gas/gas-turbines/9ha}). Last stage blades are visible on the right end of the engine.}
    \label{fig:LSB}
\end{figure}

\begin{figure}[h]
    \centering
    \includegraphics[width=0.99\textwidth]{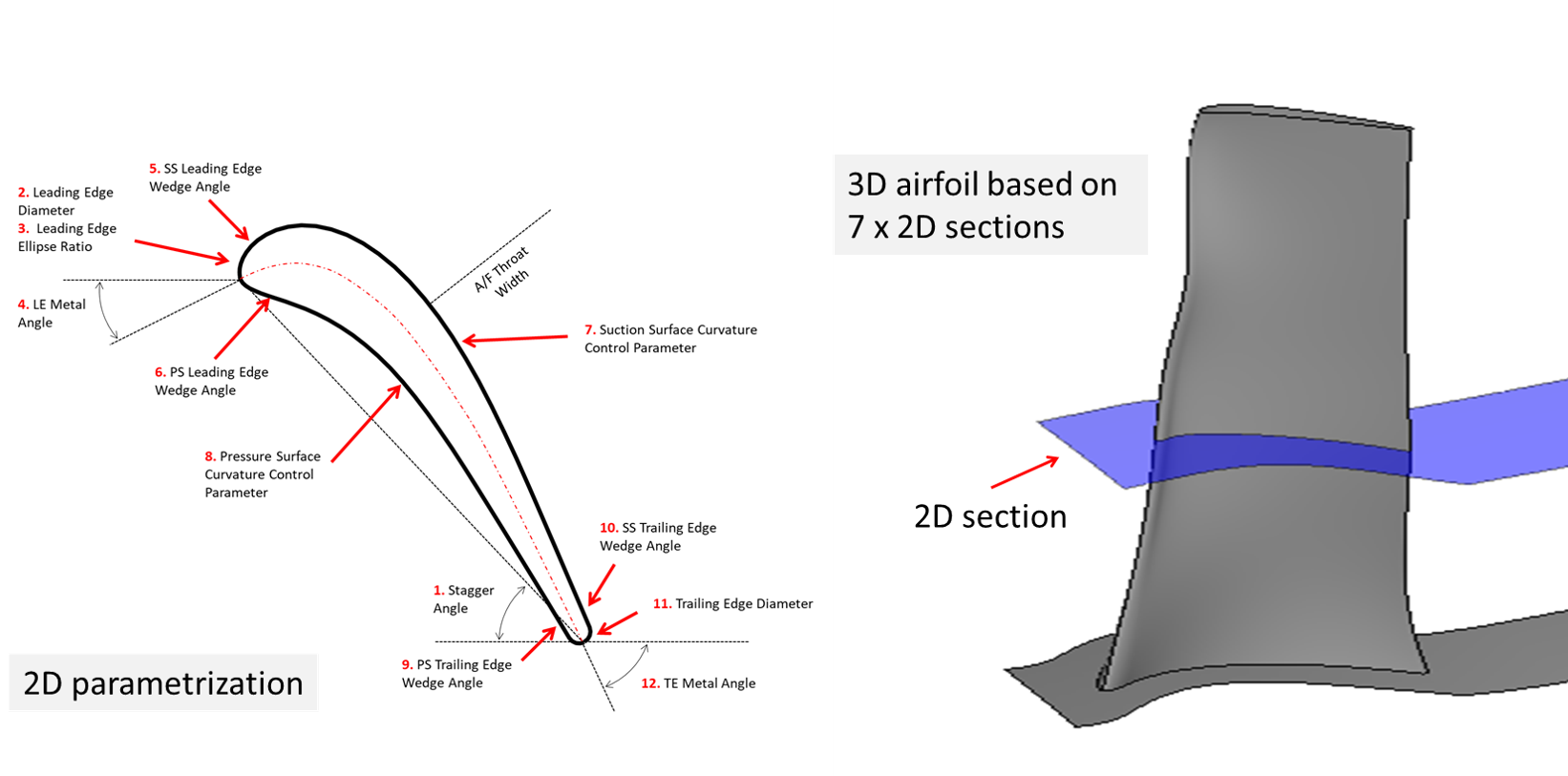}
    \caption{Last stage blade parametrization of 2D sections.}
    \label{fig:3Dairfoil_parametrization}
\end{figure}

In the current work, turbine performance is evaluated using steady-state RANS CFD at two different fidelity levels; a fast running coarse mesh for broader design space exploration, and a slower running fine mesh for accuracy refinement. The 3D airfoil is parametrized using the approach illustrated in Figure \ref{fig:3Dairfoil_parametrization}, which shows the airfoil parametrization at one section on the left, and a view of the full 3D rotor on the right.
The airfoil surface is constructed with 7 2D profiles at different spanwise locations from hub to tip. Each airfoil section is characterized by 12 independent parameters. The ranges for these parameters are selected to provide a wide design space while respecting geometrical constraints. The 12 parameters, as sketched in Figure \ref{fig:3Dairfoil_parametrization}, include the stagger angle, leading and trailing edge metal angles, leading edge diameter, suction and pressure side wedge angles, leading edge and trailing edge metal angles, and additional parameters to control the airfoil curvature between the leading and trailing edges.

The 2D sections are aligned relative to each other in circumferential and axial space by aligning the section centers of gravity (CG) along a radial line through the hub section CG (referred to as the stacking line). After the section CGs are aligned, one additional parameter, referred to as the airfoil lean angle, is applied to reorient the stacking line relative to the radial direction. Surfaces are fit through the seven stacked sections to create the full, continuous, 3D airfoil definition. Eighty five total parameters  are therefore required to define the complete 3D airfoil shape.

The parameters are expressed as offsets from a baseline, requiring that each section starts from an appropriate reference design. An in-house software package tailored specifically for turbomachinery design uses the parameters described above to create the airfoil coordinates, and these coordinates are then transformed into a full 3D CAD model of the rotor blade for the CFD grid generation. A 3D structured mesh is built using a commercially available software package. Grid templates are built from the baseline geometry and used consistently for all the cases throughout the optimization. With this approach, all the grids have similar refinement and quality metrics. To simulate the full stage, the upstream stator is included in the CFD calculation for each case.The design of the vane and the vane mesh are not altered through the optimization.

The 3D CFD analysis is performed using GE’s in-house CFD solver TACOMA, a 2nd-order accurate (in time and space), finite-volume, block-structured, compressible flow solver \cite{seeley, ren}. The steady Reynolds-Averaged Navier-Stokes (RANS) calculations are solved with a mixing plane between rotating and stationary components. Source terms are included at various locations along the endwalls to simulate the injected cooling, leakage, and purge flows.
The two objectives are to maximize the aerodynamic efficiency while matching the baseline degree of reaction. For this problem, with the full stage modeled, the degree of reaction can be calculated using standard turbine definitions.
\begin{enumerate}
\item{Aerodynamic efficiency}: The efficiency is calculated as the ratio of mechanical power and isentropic power. All inputs required to produce the efficiency value are available from the output of the CFD simulation. Design preference is to maximize efficiency.
\item{Pseudo-reaction}: In a stage calculation, the degree of reaction indicates the split of flow acceleration between stator and rotor. The degree of reaction can be calculated based on the full stage CFD, so the flow quantities between stator and rotor are extracted from the numerical results for each case. This objective is monitored to ensure that the changes in airfoil shape do not significantly affect the turbine operating condition. The design preference of this objective is to be as close as possible to the baseline value.
\end{enumerate}

\begin{figure}[h]
    \centering
    \includegraphics[width=0.65\textwidth]{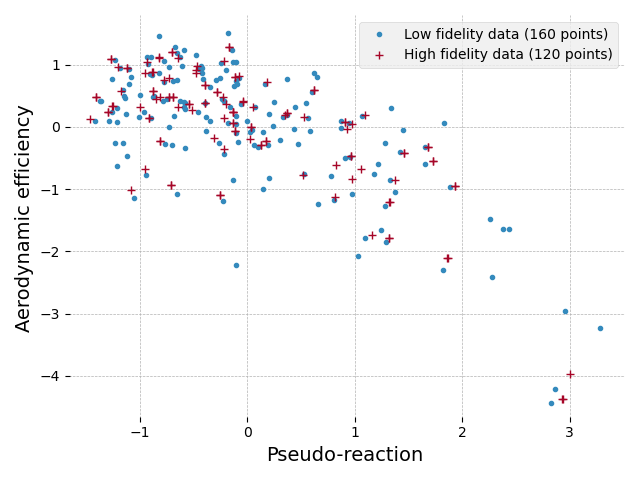}
    \caption{Airfoil shape optimization. Observed outputs from the two-fidelity solvers to be used to construct the ARGP metamodel.}
    \label{fig:3Dairfoil_DOE}
\end{figure}

Multiple quantities are extracted and processed from the CFD, like ideal Mach number at various spanwise locations. Radial profiles of flow quantities, like pressure and flow angle, are obtained to characterize the quality of flow field propagating  from the turbine to the downstream exhaust diffuser, which has not been modeled here. Additional objectives are formulated based on the desirability of the turbine exit flow profile. Diffusion rate, diffusion ratio, and shock intensity are assessed for each case to provide additional selection criteria.
The resulting simulated data consists of 160 shape configuration points are generated to be run on the coarse grid and another 120 for the fine grid that have successfully converged to CFD solutions. The two main objectives, efficiency and degree of reaction, are plotted in Figure \ref{fig:3Dairfoil_DOE}, as variations from each respective baseline. The cases assessed with a low fidelity grid are represented with gray circles, while the cases assessed with high fidelity are represented with blue diamonds. Given the small number of points compared to the number of parameters, a conclusive Pareto front has not yet been identified.

\begin{figure}[h]
    \centering
    \includegraphics[width=0.49\textwidth]{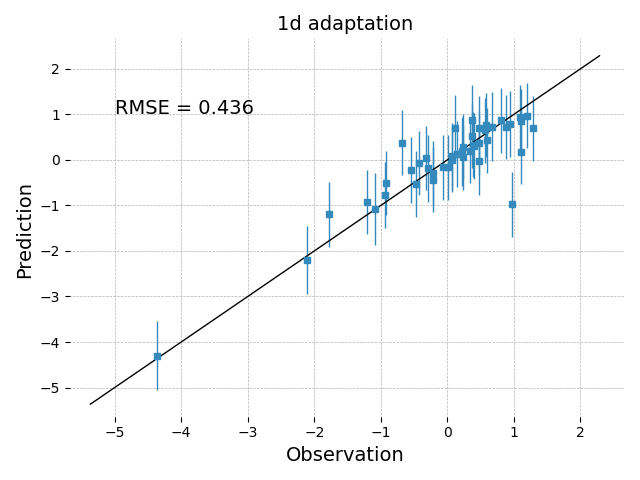}
    \includegraphics[width=0.49\textwidth]{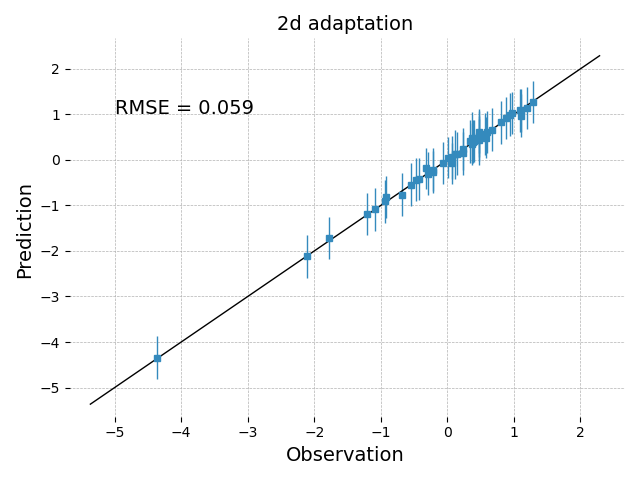}
    \includegraphics[width=0.49\textwidth]{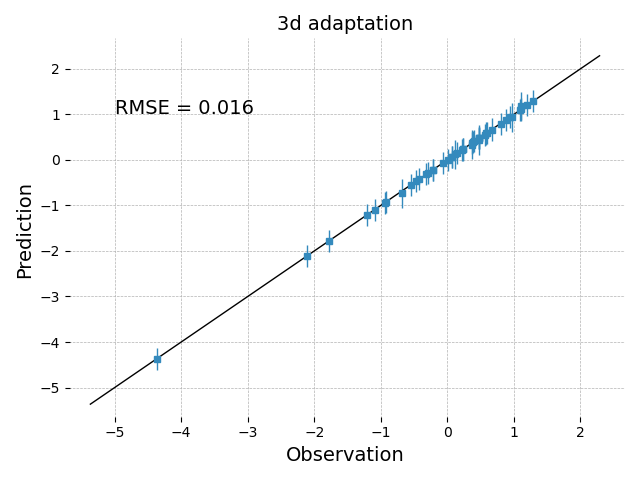}
    \caption{Airfoil shape optimization. Test observations of aerodynamic efficiency vs ARGP predictions for $d = 1, 2, 3$.}
    \label{fig:airfoil_qoi1}
\end{figure}

\begin{figure}[h]
    \centering
    \includegraphics[width=0.49\textwidth]{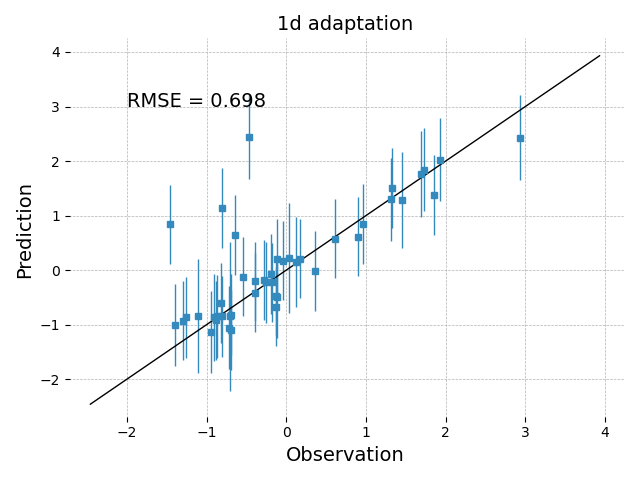}
    \includegraphics[width=0.49\textwidth]{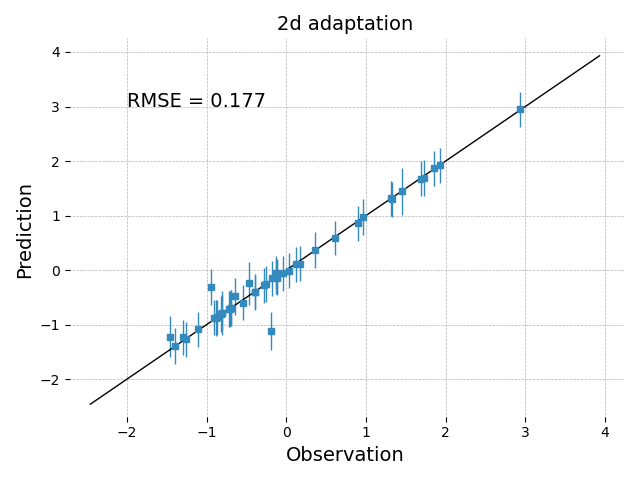}
    \includegraphics[width=0.49\textwidth]{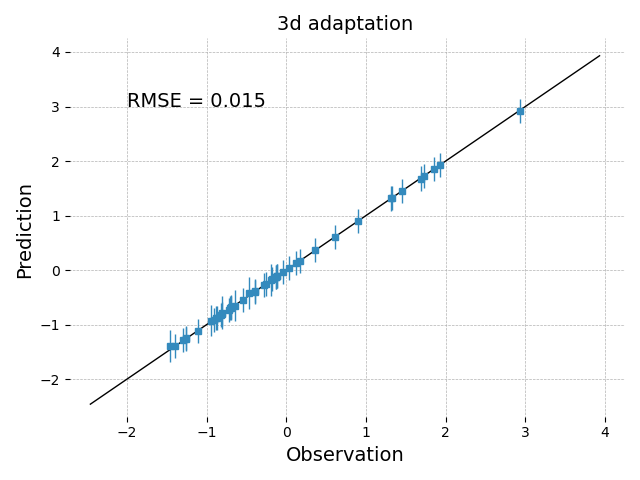}
    \caption{Airfoil shape optimization. Test observations of pseudo-reaction vs ARGP predictions for $d = 1, 2, 3$.}
    \label{fig:airfoil_qoi2}
\end{figure}

\begin{figure}
    \centering
    \includegraphics[width=0.49\textwidth]{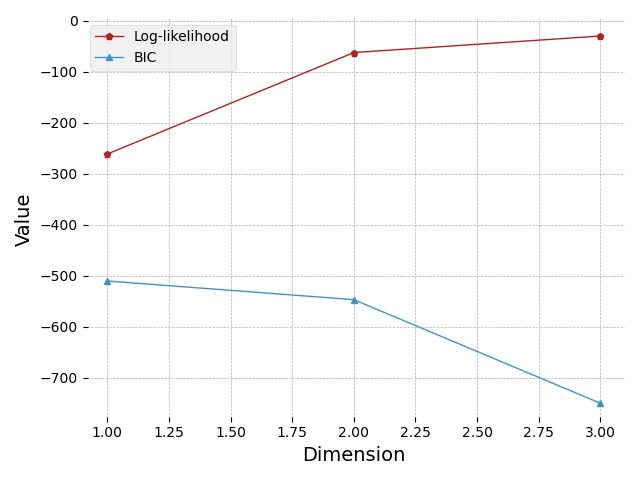}
    \includegraphics[width=0.49\textwidth]{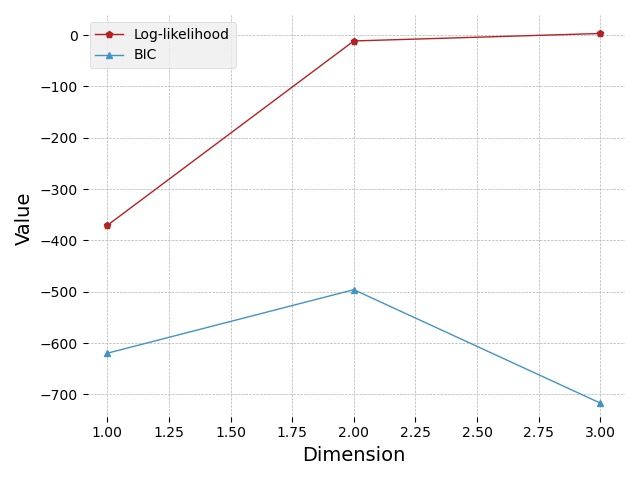}
    \caption{Airfoil shape optimization. Maximum log-likelihood and Bayesian Information criterion values for reduced dimension $d = 1, 2, 3$.}
    \label{fig:bic}
\end{figure}

For this problem, we fix the correlation coefficient between the two models to $\rho_1 = 1$ since the different level of accuracy of the two solvers varies as a result of different mesh discretization, unlike the previous examples where higher fidelity codes resulted from scaling the low fidelity ones. We split the available high fidelity observations to 75 training points and 45 test points while all low fidelity observations are used in training the ARGP using MCMC. As expected, Algorithm \ref{alg:two_step} converges slower than in the previous synthetic examples due to the higher dimensionality. Figs. \ref{fig:airfoil_qoi1} \& \ref{fig:airfoil_qoi2} show the comparison of the ARGP mean predictions ($\pm$ 2-standard deviations) for the two quantities of interest vs the $45$ test observations for reduced dimensionalities $d = 1, 2, 3$. Clearly the fit improves significantly as $d$ increases and reported residual mean square error (RMSE) value simply confirms this fact by reducing to as low as $0.016$ \& $0.015$ respectively. The quality of fit on the test data for $d=3$ can serve as a stopping criterion when testing for different reduced dimensionalities $d$ and no need for further exploration on $d\geq 4$ is required. To further support this claim we report values of the loglikelihodd function and the Bayesian Information criterion (BIC) \cite{bishop}. The latter is a standard criterion used for model selection and is given by the loglikelihood expression penalized by a term that depends on the number of training data points and model parameters. Specifically, 
\begin{equation}
    \mathrm{BIC} = -\bell(\bphi^*) - \frac{1}{2} \vert\bphi\vert \log \sum_{t=1}^s n_t,
\end{equation}
where $\vert \bphi \vert$ is the number of model parameters to be inferred and $\bphi^*$ is the maximum likelihood estimate. The largest values of BIC typically indicate the most favorable model.  We plot the maximum log-likelihood and the BIC values as a function of $d$ in Fig. \ref{fig:bic}. We observe that the log-likelihood becomes almost constant, indicating that no further improvement in terms of fitting the training set can be achieved. The BIC on the other hand drops at $d=3$, indicating that the number of parameters (penalty term) becomes significantly large, thus the trade-off between data-fit improvement and number of parameters to be inferred has already a negative trend.

\begin{figure}[h]
    \centering
    \includegraphics[width=0.8\textwidth]{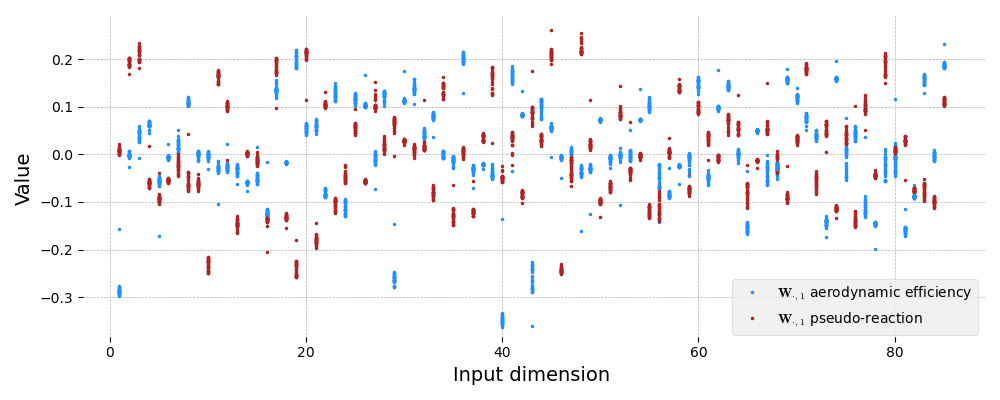}
    \caption{Airfoil shape optimization. Entries of the 1st column $\bW_{\cdot, 1}$ of the projection matrix $\bW$ corresponding to the two QoIs.}
    \label{fig:airfoil_w}
\end{figure}

At last, the display the sample values of the 1st column of $\bW$, denoted as $\bW_{\cdot, 1}$ for both QoIs in Fig. \ref{fig:airfoil_w}. The values corresponding to the same input dimension in general do not seem to be similar, indicating that the importance of each input parameter on the different QoIs varies. Most importantly, very few of the $\bW{\cdot, 1}$ entries are near zero, thus only a small number of the parameters in the original input space are negligible with no effect in the output QoIs. This allows us to conclude that the reduction from $D=85$ to $d = 3$ is not the result of simply discarding unimportant parameters but that our algorithm was able to reveal the low-dimensional linear embedding that captures the the models observed behavior accurately. 

\section{Conclusions}

We have thus far presented a methodology to exploit low dimensional subspaces for input dimensionality reduction and have demonstrated it's applicability to uncertainty propagation problems using multi-fidelity simulations. This was achieved by employing an autoregressive Gaussian Process scheme with a linear embedding that is modeled using an orthogonal matrix that maps the original input variable to its latent counterpart. We treat the matrix as an additional set of hyperparameters and we learn the mapping jointly with training the ARGP by employing a two-step algorithm that train the ARGP using standard MLE and MCMC techniques while it updates the orthogonal matrix using the proposed Geodesic Monte Carlo sampling. We validated our method on synthetic examples with known linear embedding and we further utilised it to study the problem of 3D airfoil optimization by building a low dimensional responce surface to understand the impact of airfoil shape parameters on the particular quantities of interest.

Our work continues on the findings of \cite{constantine_AS, tripathy_bilionis} towards the development of a surrogate that enables a fully Bayesian exploration of the active subspace. Quoting Tripathy \& Bilionis \cite{tripathy_bilionis} ``the big challenge is the construction of proposals that force $\bW$ to remain on the Stiefel manifold $\dots$ such approaches would open the way for more robust AS dimensionality selection". We have achieved a Bayesian treatment that relieves us from such headaches by tailoring our HMC sampling method \cite{byrne} particularly on the Stiefel manifold. In addition, the capability to leverage information from computer codes of a varying degree of accuracy and cost is another novel step that opens new directions in multi-fidelity simulations for uncertainty quantification, including multi-output Gaussian Processes and multi-objective Bayesian optimization. Such topics are to be explored in future research. 

\section*{Acknowledgments}

The information, data, or work presented herein was funded in part by the Advanced Research Projects Agency-Energy (ARPA-E), U.S. Department of Energy, under Award Number DE-AR0001204. The views and opinions of authors expressed herein do not necessarily state or reflect those of the United States Government or any agency thereof.

\appendix

\section{Mathematical background on Geodesic Monte Carlo}
\label{sec:geod_mc_math}

\subsection{Hamiltonian dynamics on an embedded manifold $\calM$}

Consider the problem of sampling from a probability density $p(x)$ defined on a manifold $\calM$ that is \emph{isometrically} embedded in $R^D$ and equipped with a metric tensor $G(x)$, serving as the isometry between the two spaces, thus, preserving the distance. In order to employ a Hamiltonian Monte Carlo algorithm \cite{neal}, the Hamiltonian function is formed 
\begin{equation}
    H(x, v) = -\log p(x) + \frac{1}{2}v^T G^{-1}(x) v
\end{equation}
and its dynamics are described by 
\begin{eqnarray}
\left\{ \begin{array}{lrl} \dot{x} =
     & \displaystyle{\frac{\partial H}{\partial v} } = & G(x)^{-1}u \\ \dot{v} =
     & \displaystyle{\frac{\partial H}{\partial x} } = & \nabla_x\left[\log p(x) - \frac{1}{2}v^T G^{-1}(x)v\right] 
\end{array} \right.
\end{eqnarray}
Integrating the Hamiltonian on a Euclidean space typically involves a \emph{Stormer-Verlet leapfrog scheme} \cite{bishop, hairer} which in our case requires inverting the metric tensor $G(x)$. To avoid numerical drawbacks associated with such operations, Byrne \& Girolami \cite{byrne} suggest splitting $H(x,v)$ to two distinct Hamiltonians, namely $H^{[1]}(x, v) = -\log p(x)$ with dynamics 
\begin{eqnarray}
\begin{array}{lrl} \dot{x} = & \displaystyle{\frac{\partial H^{[1]}}{\partial v}} = & 0 \\
    \dot{v} = & -\displaystyle{\frac{\partial H^{[1]}}{\partial x}} = & \nabla_x \log p(x)
\end{array}
\end{eqnarray}
and $H^{[2]}(x, v) = \frac{1}{2}v^T G^{-1}(x)v$ whose dynamics are described as a geodesic flow over a geodesic curve $\gamma(t)$ that maintains a constant velocity $\vert\vert\dot{\gamma}(t)\vert\vert_G$ and is known as \emph{Levi-Civita connection} of $G$ \cite{abraham}.

\subsection{Hamiltonian dynamics on the Stiefel manifold $\calV_{d, D}$}

For the special case where $\calM = \calV_{d, D}$, with target density function being $p(\bw\vert \bphi, \calZ)$, the Hamiltonian is given by eq. (\ref{eq:hamiltonian}). The tangent space at a point $\bw\in \calV_{d, D}$ is given by 
\begin{equation}
    T_{\bw}\calV_{d, D}=  \left\{ \bZ \in \R^{D\times d} : \bZ^T\bw + \bw^T\bZ = 0\right\}
\end{equation}
and the projection of a vector $\bu$ on $T_{\bw}\calV_{d, D}$ is given by 
\begin{equation}
    \Pi_{\bw}\bu = \bu - \frac{1}{2}\bw\left( \bw^T \bu + \bu^T \bw \right) .
\end{equation}
Then one can see that for the Hamiltonian defined in eq. (\ref{eq:hamiltonian}), the dynamics of $H^{[1]}$ imply that $\bw(t)$ is constant everywhere, while $\bu(t)$ satisfies 
\begin{equation}
    \bu(t) = \bu(0) + t\Pi_{\bw_0} \nabla_{\bw} \log p(\bw\vert \bphi, \calZ)\big\vert_{\bw = \bw(0)}.
\end{equation}
On the other hand, the dynamics of $H^{[2]}(\bw, \bu)$ follow the geodesic flow given as \cite{byrne}
\begin{equation}
\label{eq:geod_nd}
    \left[\bw(t)\ \bu(t)\right] = \left[\bw(0)\ \bu(0) \right]\exp\left\{t \left[\begin{array}{cc}
        A & -S(0) \\
        I & A
    \end{array}\right]\right\}\left[\begin{array}{cc}
        \exp\left\{-tA\right\} & 0 \\
        0 & \exp\left\{-t A\right\}
    \end{array}\right].
\end{equation}
In the above, $A := \bw^T\bu(t)$ is constant over the geodesic and $S(t) = \bu^T(t)\bu(t)$. For the case where $d=1$, that is $\calV_{1, D}$ is the hypersphere $\calS^{D-1}$, the geodesic flow simplifies to 
\begin{equation}
\label{eq:geod_1d}
    \left[\bw(t)\ \bu(t)\right] = \left[\bw(0)\ \bu(0) \right] \left[\begin{array}{cl}
        1 & 0 \\
        0 & \beta^{-1}
    \end{array}\right] \left[\begin{array}{cr}
        \cos(\beta t) & -\sin(\beta t) \\
        \sin(\beta t) & \cos(\beta t)
    \end{array}\right]\left[\begin{array}{cc}
        1 & 0 \\
        0 & \beta
    \end{array}\right].
\end{equation}
Here, $\beta := \vert\vert\bu(t)\vert\vert$ is the constant angular velocity. To initialize the Hamiltonian Monte Carlo sampling algorithm, the initial momentum $v_0$ in the arbitrary manifold case is sampled from a Gaussian $\calN (0, G(x_0))$ which for $\calV_{d, D}$ becomes $\bu_0 \sim \calN(0, \Pi_{\bw_0})$. In practice we sample $u \sim \calN(0, I_{D, d})$ and set $\bu_0 = \Pi_{\bw_0}(u)$.

\section{The Matrix-Langevin distribution}
\label{sec:mL}

The Matrix-Langevin, or von Mises-Fisher distribution defined on the Stiefel Manifold $\calV_{d, D}$ has probability density given by 
\begin{equation}
    p(\bw) = \frac{1}{c(\bF)}\exp\left\{\textrm{Tr}\left[\bF^T \bw\right]\right\},
\end{equation}
where the normalizing constant is $c(\bF) = {}_0F_1(\frac{1}{2}D, \frac{1}{4}\bF^T\bF)$ and ${}_0F_1(\cdot, \cdot)$ is the hypergeometric constant with matrix arguments. It is often a convenient practice to parametrize the density function using a singular value decomposition (SVD) of the matrix $\bF \in \R^{D\times d}$ such that $\bF = \bU \bSigma \bV^T$, where $\bU$ and $\bV$ are in $\calV_{d,d}$ and $\calV_{D, D}$ respectively and $\bSigma$ is the $d\times D$ diagonal matrix that contains the singular values of $\bF$. In this case, the normalizing constant simplifies to $c(\bF) = {}_0F_1(\frac{1}{2}D, \frac{1}{4}\bSigma^2)$ and the mode of the distribution is given by $\bU\bV^T$ \cite{khatri}. This intuitively means that $\bU$ and $\bV$ are orientation matrices that determine the directions where $\bW$ is concentrated while the diagonal entries of $\bSigma$ control the level of concentration in those directions. Specifically, when $\textrm{diag}(\bSigma) \to 0$, the concentration becomes very broad and eventually converges to the uniform probability measure on the Stiefel manifold, while $\textrm{diag}(\bSigma) \to \infty$ results in Dirac measures on the directions specified by $\bU$ and $\bV$. To simplify things further, in our implementations we consider $\bV = \bI_{D}$, therefore $\bF = \bU \bSigma$ and the mode becomes $\bU$ with the columns of $\bW$ being thus, centered around the columns of $\bU$. An illustrative configuration of $\bF$ on $\calV_{2,3}$ is shown in \cite{tsilifis_rspa}. In our implementations, we take advantage of these features in order to adjust $\bSigma$ accordingly so as to specify very broad or tight priors for the columns of $\bW$ to be centered around orientations specified in $\bU$ that become available from lower dimensional adaptations. 

\section*{}
\bibliographystyle{plain}
\bibliography{ref}

\end{document}